\pdfoutput=1
\documentclass[11pt]{article}
\usepackage{booktabs}
\usepackage{multirow}
\usepackage{subfigure}
\usepackage{subcaption}
\usepackage{ACL2023}
\usepackage[utf8]{inputenc}

\usepackage[arabic,english]{babel} % Support for Arabic and English languages
\usepackage{array}
\usepackage{appendix}
\usepackage[printonlyused]{acronym}
\usepackage{hyperref} % Load the hyperref package for hyperlinks
\usepackage{longtable}
\usepackage{geometry}
\usepackage{times}
\usepackage{latexsym}
\usepackage{graphicx}
\usepackage{adjustbox}
\usepackage{float}
\usepackage{nicematrix}
\usepackage{siunitx}
\usepackage{bm}
\usepackage{soul}
\usepackage[T1]{fontenc}
\usepackage{microtype}

\usepackage{xcolor}
\usepackage{colortbl}
\usepackage{xspace}
\usepackage{ifthen}

\usepackage{inconsolata}
\usepackage{cleveref}
\usepackage{booktabs}

\usepackage{amssymb}% http://ctan.org/pkg/amssymb
\usepackage{pifont}% http://ctan.org/pkg/pifont
\usepackage{amsmath}

\title{Comparison of Multilingual and Bilingual Models for Satirical News Detection of Arabic and English}

 \author{Omar W. Abdalla \\
   University of New South Wales, Sydney \\
   \texttt{o.abdalla@student.unsw.edu.au} \\ \And
   Aditya Joshi \\
   University of New South Wales, Sydney \\
   \texttt{aditya.joshi@unsw.edu.au} \\ \AND
   Rahat Masood \\
   University of New South Wales, Sydney \\
   \texttt{rahat.masood@unsw.edu.au} \\ \And
   Salil S. Kanhere \\
   University of New South Wales, Sydney \\
   \texttt{salil.kanhere@unsw.edu.au} \\ }

\begin{document}
\maketitle
\begin{abstract}
    Satirical news is real news combined with a humorous comment or exaggerated content, and it often mimics the format and style of real news. However, satirical news is often misunderstood as misinformation, especially by individuals from different cultural and social backgrounds. This research addresses the challenge of distinguishing satire from truthful news by leveraging multilingual satire detection methods in English and Arabic. We explore both zero-shot and chain-of-thought (CoT) prompting using two language models, Jais-chat(13B) and LLaMA-2-chat(7B). Our results show that CoT prompting offers a significant advantage for the Jais-chat model over the LLaMA-2-chat model. Specifically, Jais-chat achieved the best performance, with an F1-score of 80\% in English when using CoT prompting. These results highlight the importance of structured reasoning in CoT, which enhances contextual understanding and is vital for complex tasks like satire detection.
\end{abstract}

% \textcolor{red}{need to update with more insightful results.} 
% \end{abstract}

% \textcolor{red}{Line 26: The in-text citation format seems incorrect. ``Satire is the act of making fun of someone or something intending to embarrass or discredit them (Assiri and Himdi, 2023)(Burfoot and Baldwin, 2009).''}
% \textcolor{blue}{In the first citation, the exact words of the researchers that are relevant are: ``Satirical news is defined as the use of humor, irony, exaggeration, or ridicule to expose and criticize certain viewpoints.'' As for the second citation, they define satire as: `` Satire is a form of communication where humor and irony are used to criticize someone's behavior and ridicule it.''}

\section{Introduction}
Satire is the act of making fun of someone or something intending to embarrass or discredit them~\cite{app131910616}\cite{burfoot2009automatic}. 
% It is usually used for humorous purposes, however, this often reflects some form of criticism. 
Satire is context-dependent, which is why satirical news can sometimes be mistaken for misinformation, even though there is no intention of misleading any parties, making satirical news prone to being misclassified as ``false positive'' misinformation~\cite{levi-etal-2019-identifying}.

Most existing methods focus on satire detection in a single language, with limited research on multilingual approaches. Zero-shot prompting of large language models (LLMs) has been explored, but this technique struggles with satire detection due to a lack of context. This research investigates how CoT prompting improves the accuracy of bilingual and multilingual models, using Jais-chat~\cite{sengupta2023jaisjaischatarabiccentricfoundation}\footnote{Jais-chat has been reported as a bilingual Arabic-English model.} and LLaMA-2-chat~\cite{touvron2023llama2openfoundation}. Bilingual models like Jais-chat are trained on only two languages, English and Arabic in our case, while multilingual models like LLaMA-2-chat are trained on more than two languages. Our paper provides insight into how specialized, language-focused training compares to more general, multilingual training, particularly in the context of satire detection for English and Arabic texts.

This research aims to answer: \textit{i) How does the performance of a bilingual model compare to a multilingual model in detecting satire across languages?} and \textit{ii) What impact does CoT prompting have on accuracy?} We evaluate Jais-chat\footnote{\href{https://huggingface.co/inceptionai/jais-13b-chat}{https://huggingface.co/inceptionai/jais-13b-chat}} and LLaMA-2-chat\footnote{\href{https://huggingface.co/meta-llama/Llama-2-7b-chat-hf}{https://huggingface.co/meta-llama/Llama-2-7b-chat-hf}} across two languages (English and Arabic) using CoT prompting. Our results indicate that CoT prompting outperforms zero-shot prompting for satire detection, particularly with the Jais-chat model, whereas LLaMA-2-chat showed minimal improvements with CoT, maintaining consistent performance across both prompting methods. Our contributions include:
\begin{itemize}
    \item We study and apply  Chain-of-Thought (CoT) prompting for satire detection in both English and Arabic, guiding the model through a step-by-step reasoning process for improved accuracy. %Satire Detection using CoT: To the best of our knowledge, our study is first to propose the use of CoT prompting for satire detection in both English and Arabic languages. Our contribution leverages CoT prompting to guide the model through a step-by-step reasoning process, allowing it to break down the problem into smaller, more manageable parts rather than directly generating an answer. This encourages deeper analysis and improves the model's ability to accurately detect satire.
    \item We introduce multilingual prompting for satire detection, tackling challenges related to cultural nuances and different humor styles across the two languages, English and Arabic. %Multilingual Satire Detection: To the best of our knowledge, our study is the first to propose using multilingual prompting for satire detection. Handling satire across multiple languages introduces unique challenges, such as understanding cultural nuances and different styles of humor. %We prompt LLMs to detect satire in two different languages, and conduct experiments using two different models: Jais-chat and LLaMA-2-chat, across four different datasets i.e.,  two English datasets and two Arabic datasets.
    \item We compare the performance of a bilingual model against a multilingual model, providing insights into their effectiveness in satire detection across different languages.
    % \item Best Results: Our results show that CoT performs better in satire detection than zero-shot prompting, especially with the Jais-chat model, which achieved the highest performance with an F1-score of 80\% when using CoT prompting with English prompts, and an F1-score of 70\% with Arabic prompts. \textcolor{blue}{In comparison, LLaMA-2-chat showed minimal improvements when using CoT, featuring its consistency when prompted using zero-shot and CoT} %zero-shot prompting resulted in lower F1-scores, with 67\% for English datasets and 64\% for Arabic datasets, respectively. These results demonstrate the effectiveness of CoT prompting in enhancing the model's ability to detect satire, particularly in multilingual contexts.
\end{itemize}

The rest of the paper is organized as follows: Section \ref{sec:related} reviews the prior research on satire detection. Section \ref{sec:methodology} outlines our proposed methodology and experiment setup. Section \ref{sec:results} presents the results of our experiments, and Section \ref{sec:conclusion} concludes the paper with a discussion of findings and future work.

% There are many news websites today.
% Some are satire. What is satire?
% Relationship between satire and fake news.
% Arabic and English.
% Jais-chat.

\begin{figure*}[ht!]
  \centering
  \includegraphics[width=\textwidth, keepaspectratio]{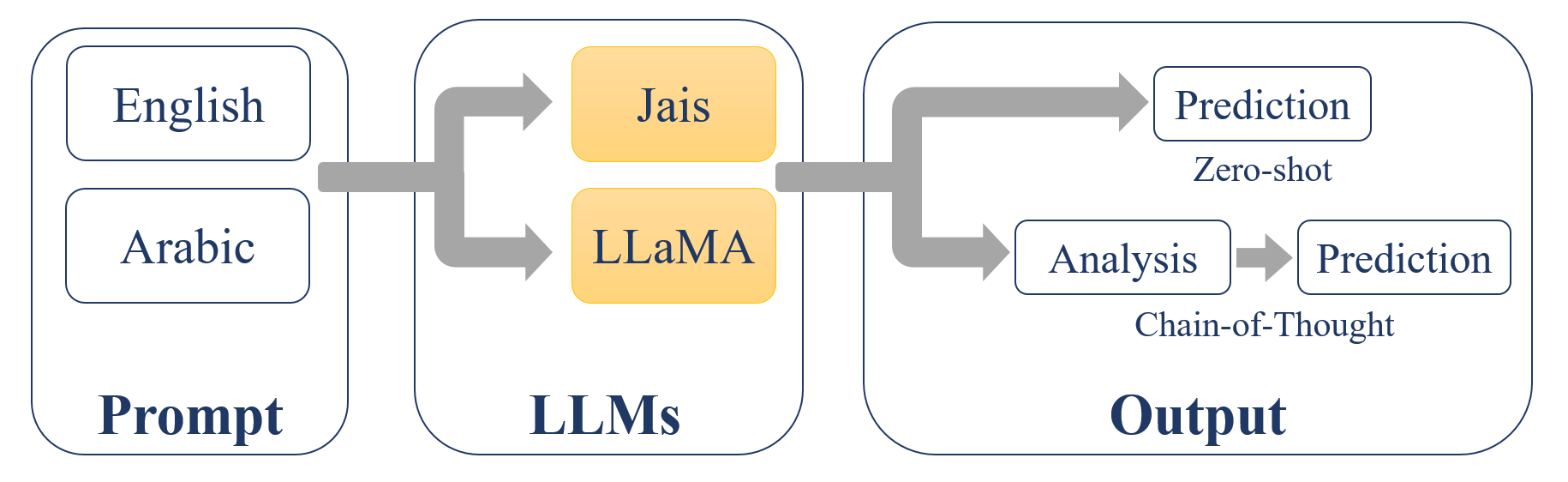}
  \caption{Overview of Methodology.}
  \label{fig:zero_cot_process}
\end{figure*}

\section{Related Work}\label{sec:related}
% Different approaches to satire detection have been explored, utilizing different languages and datasets. Barbieri et al. (2015)\cite{barbieri2015tweet} examined Spanish tweets, comparing models based on lexical and semantic features with a simpler bag-of-words (BoW) approach, finding that their method performed better. McHardy et al. (2019)\cite{mchardy-etal-2019-adversarial} tackled satire in German news by using adversarial training to ensure the model focused on content rather than publication-specific details, improving results. Rogoz et al. (2021)\cite{rogoz-etal-2021-saroco} introduced the SaRoCo dataset for Romanian satire detection, where Ro-BERT outperformed CNN models. However, it still fell short of human-level accuracy, highlighting the complexity of the task.

Satire detection methods have progressed from basic lexical and semantic features, such as bag-of-words (BoW) models and handcrafted features like frequency, sentiment, and part-of-speech (POS) tags~\cite{barbieri2015tweet,burfoot2009automatic,frain-wubben-2016-satiriclr}, to advanced machine learning and deep learning approaches. Earlier methods used support vector machines (SVM) and semantic checks for coherence in named entities~\cite{burfoot2009automatic}, while more recent techniques incorporate attention mechanisms, adversarial training, and transformers like BERT and GPT~\cite{mchardy-etal-2019-adversarial, rogoz-etal-2021-saroco, saadany2020fake, app131910616}. Some studies have explored multimodal methods, integrating both text and images, with models like ViLBERT excelling in this area~\cite{li-etal-2020-multi-modal}. In Arabic satire detection, CNNs and linguistic markers, such as sentiment and first-person pronouns, have proven effective, while transformers have also shown strong performance~\cite{saadany2020fake, app131910616}.

% The rapid development of LLMs has raised significant concerns about their ability to produce convincing misinformation. Studies by Chen and Shu (2023)\cite{chen2023can} and Zhou et al. (2023)\cite{zhou2023synthetic} have shown that AI-generated content can be tricky to identify, even for sophisticated systems. Wan et al. (2024)\cite{wan2024dell} developed the DELL framework to improve misinformation detection by combining expert insights and diverse methods, while Lai et al. (2024)\cite{lai2024adaptive} used adaptive ensemble learning to boost detection accuracy. As LLMs become more advanced, researchers like Sun et al. (2024)\cite{sun2024exploring} and Satapara et al. (2024)\cite{satapara2024fighting} are working on new techniques to stay ahead of deceptive AI, stressing the need for public awareness and better detection strategies.

% Previous research has demonstrated improvements in satire detection over time. However, challenges remain, particularly with multilingual support and the application of CoT prompting in satire detection. In this paper, we address these limitations by leveraging the capabilities of two powerful models: Jais-chat and LLaMA-2-chat, both trained on English and Arabic. By integrating these models with the well-established benefits of the CoT approach, we aim to achieve more accurate and nuanced satire detection outcomes.
Despite advancements in satire detection, challenges persist, especially with multilingual support and CoT prompting. This paper tackles these issues by leveraging the Jais-chat and LLaMA-2-chat models, both trained on English and Arabic, and integrating them with CoT to enhance accuracy and nuance in satire detection.
% \textcolor{red}{a paragraph on limitations of existing work and your unique contributions. }

% Key papers in satire detection.
% Satire detection for Arabic.
% Satire detection for English.
% Misinformation

\section{Methodology}\label{sec:methodology}

\subsection{Overview}
We apply zero-shot prompting to the selected datasets and compare their performance against CoT prompting. Zero-shot prompting instructs the model to perform a task without providing any examples for guidance, whereas CoT prompting involves appending instructions such as ``Describe your reasoning in steps'' or ``Explain your answer step by step'' to the query, encouraging the model to think through the problem before responding. As illustrated in Figure~\ref{fig:zero_cot_process}, we use prompts in English and Arabic with two models, Jais-chat and LLaMA-2-chat, to generate outputs based on the input prompts. 
% \textcolor{red}{This is a repetition. Can we please rephrase? The focus of the study is on the comparison between a bilingual and a multilingual model in the scope of the two languages that the bilingual model supports to see and compare the performance of these models and the effect of the data they are trained on, hence the focus on only English and Arabic text.} 
To assess model robustness, we employ a multilingual prompting strategy, testing four prompt configurations: an English pre-prompt with English article text, an English pre-prompt with Arabic article text, an Arabic pre-prompt with English article text, and an Arabic pre-prompt with Arabic article text. This approach allows us to evaluate the impact of aligning the prompt language with the article language, as well as to analyze the effect of each language independently on model performance in satire detection. We assess the performance of the models by prompting them to make direct predictions (zero-shot) and compare these results with those obtained when prompting the models to first analyze the articles and then classify them based on this analysis (CoT).

We employed different prompts for zero-shot and CoT tasks. For example, the English prompt for the zero-shot is: ``\textit{You will be provided with a news article, and you are required to determine (predict) whether the article is satirical or not. Your answer should only be ``1'' if the article is satirical or ``0'' if the article is serious. Do not provide any explanation or additional commentary. Do not answer with blank.}'' 
For CoT, two prompts are used. One for the analysis phase and another one for the prediction phase. 
% The analysis prompt is: ``\textit{You will be provided with a news article, and you are required to explain why this article might be satirical or not satirical. Your explanation needs to be logical}'', whereas the prediction prompt is ``\textit{You will be provided with an analysis about a news article, and you are required to determine (predict) whether the analysis points that the article is satirical or not. Your answer should only be ``1'' if the article is satirical or ``0'' if the article is serious. Do not provide any explanation or additional commentary. Do not answer with blank.}'' 
All prompts were written in English and Arabic to assess the models' multilingual capabilities.

% Diagram essential.
% Prompt in English versus Prompt in Arabic.
% Explanation versus none.

\subsection{Data Statistics}\label{sec:experiment}

\begin{table*}[ht!]
\centering
\caption{Summary statistics of the datasets}
\begin{tabular}{ccccc}
\toprule
\textbf{Attribute} & \textbf{Assiri} & \textbf{Saadany} & \textbf{Phosseini} & \textbf{SatiricLR} \\ \midrule
Language & Arabic & Arabic & English & English \\ 
Number of Entries & 1525 & 7948 & 6943 & 3411 \\ 
Average Words per Article & 1013 & 1635 & 2721 & 2472 \\ 
Satire (\%) & 760 (49.8\%) & 3185 (40\%) & 3956 (57\%) & 1706 (50\%) \\ 
Non-Satire (\%) & 765 (50.2\%) & 4763 (60\%) & 2987 (43\%) & 1705 (50\%) \\ \bottomrule
\end{tabular}
\label{tab:dataset_stats}
\end{table*}

\begin{table*}[ht!]
\centering
\caption{Performance of Jais-chat and LLaMA-2-chat Models on Different Datasets and Languages}
\begin{tabular}{ccccccccc}
\toprule
\multirow{3}{*}{Model} & \multirow{3}{*}{Prompt} & \multirow{3}{*}{Dataset} & \multirow{3}{*}{Approach} & \multicolumn{4}{c}{Performance Metrics} \\ \cline{5-8} 
                       &                           &                          &                           & Accuracy & Precision & Recall & F1-Score \\ \hline
                       
\multirow{16}{*}{Jais} & \multirow{8}{*}{English}  & \multirow{2}{*}{Assiri} & Zero-shot                 & 65.6     & 72.7      & 49.9   & \textit{59.2}     \\ \cline{4-8} 
                       &                           &                          & Chain-of-Thought           & 79.9     & 79.7      & 80.0   & \textbf{\textit{80.0}}     \\ \cline{3-8} 
                       &                           & \multirow{2}{*}{Saadany} & Zero-shot                 & 45.6     & 18.0      & 10.1   & \textit{12.9}     \\ \cline{4-8} 
                       &                           &                          & Chain-of-Thought           & 71.5     & 62.8      & 71.2   & \textit{66.7}     \\ \cline{3-8} 
                       &                           & \multirow{2}{*}{Phosseini} & Zero-shot                 & 37.9     & 42.8      & 26.9   & \textit{33.0}     \\ \cline{4-8} 
                       &                           &                          & Chain-of-Thought           & 62.1     & 69.5      & 59.8   & \textit{64.3}     \\ \cline{3-8} 
                       &                           & \multirow{2}{*}{SatiricLR} & Zero-shot                 & 45.5     & 38.5      & 15.0   & \textit{21.6}     \\ \cline{4-8} 
                       &                           &                          & Chain-of-Thought           & 62.9     & 64.0      & 59.0   & \textit{61.4}     \\ \cline{2-8} 
                       
                       & \multirow{8}{*}{Arabic}   & \multirow{2}{*}{Assiri} & Zero-shot                 & 69.1     & 77.3      & 54.0   & \textit{63.6}     \\ \cline{4-8} 
                       &                           &                          & Chain-of-Thought           & 53.9     & 52.1      & 95.8   & \textit{67.5}     \\ \cline{3-8} 
                       &                           & \multirow{2}{*}{Saadany} & Zero-shot                 & 36.7     & 32.3      & 52.8   & \textit{40.1}     \\ \cline{4-8} 
                       &                           &                          & Chain-of-Thought           & 50.8     & 44.3      & 88.9   & \textit{59.1}     \\ \cline{3-8} 
                       &                           & \multirow{2}{*}{Phosseini} & Zero-shot                 & 57.9     & 60.4      & 75.6   & \textit{67.2}     \\ \cline{4-8} 
                       &                           &                          & Chain-of-Thought           & 60.8     & 62.4      & 78.6   & \textbf{\textit{70.0}}     \\ \cline{3-8} 
                       &                           & \multirow{2}{*}{SatiricLR} & Zero-shot                 & 46.6     & 46.6      & 46.0   & \textit{46.3}     \\ \cline{4-8} 
                       &                           &                          & Chain-of-Thought           & 58.6     & 56.4      & 76.5   & \textit{64.9}     \\ \hline
                       
\multirow{16}{*}{LLaMA} & \multirow{8}{*}{English} & \multirow{2}{*}{Assiri} & Zero-shot                 & 49.7     & 49.8      & 99.2   & \textit{66.3}     \\ \cline{4-8} 
                       &                           &                          & Chain-of-Thought           & 50.2     & 50.1      & 97.8   & \textit{66.3}     \\ \cline{3-8} 
                       &                           & \multirow{2}{*}{Saadany} & Zero-shot                 & 39.0     & 39.4      & 97.3   & \textit{56.1}     \\ \cline{4-8} 
                       &                           &                          & Chain-of-Thought           & 40.0     & 39.9      & 98.5   & \textit{56.8}     \\ \cline{3-8} 
                       &                           & \multirow{2}{*}{Phosseini} & Zero-shot                 & 56.9     & 56.9      & 99.8   & \textbf{\textit{72.5}}     \\ \cline{4-8} 
                       &                           &                          & Chain-of-Thought           & 56.9     & 57.0      & 98.8   & \textit{72.3}     \\ \cline{3-8} 
                       &                           & \multirow{2}{*}{SatiricLR} & Zero-shot                 & 50.0     & 50.0      & 100.0   & \textit{66.7}     \\ \cline{4-8} 
                       &                           &                          & Chain-of-Thought           & 50.0     & 50.0      & 99.2   & \textit{66.5}     \\ \cline{2-8} 
                       
                       & \multirow{8}{*}{Arabic}   & \multirow{2}{*}{Assiri} & Zero-shot                 & 49.9     & 49.9      & 100.0   & \textit{66.6}     \\ \cline{4-8} 
                       &                           &                          & Chain-of-Thought           & 50.2     & 50.0      & 100.0   & \textit{66.7}     \\ \cline{3-8} 
                       &                           & \multirow{2}{*}{Saadany} & Zero-shot                 & 40.1     & 40.1      & 100.0   & \textit{57.2}     \\ \cline{4-8} 
                       &                           &                          & Chain-of-Thought           & 40.2     & 40.1      & 99.8   & \textit{57.2}     \\ \cline{3-8} 
                       &                           & \multirow{2}{*}{Phosseini} & Zero-shot                 & 57.0     & 57.0      & 100.0   & \textbf{\textit{73.0}}     \\ \cline{4-8} 
                       &                           &                          & Chain-of-Thought           & 56.9     & 57.0      & 99.9   & \textbf{\textit{73.0}}     \\ \cline{3-8} 
                       &                           & \multirow{2}{*}{SatiricLR} & Zero-shot                 & 50.0     & 50.0      & 99.9   & \textit{66.6}     \\ \cline{4-8} 
                       &                           &                          & Chain-of-Thought           & 50.3     & 50.1      & 100.0   & \textit{66.8}    \\ \bottomrule

\end{tabular}

\label{tab:results}
\end{table*}

\begin{figure*}[ht!]
  \centering
  \includegraphics[width=\textwidth]{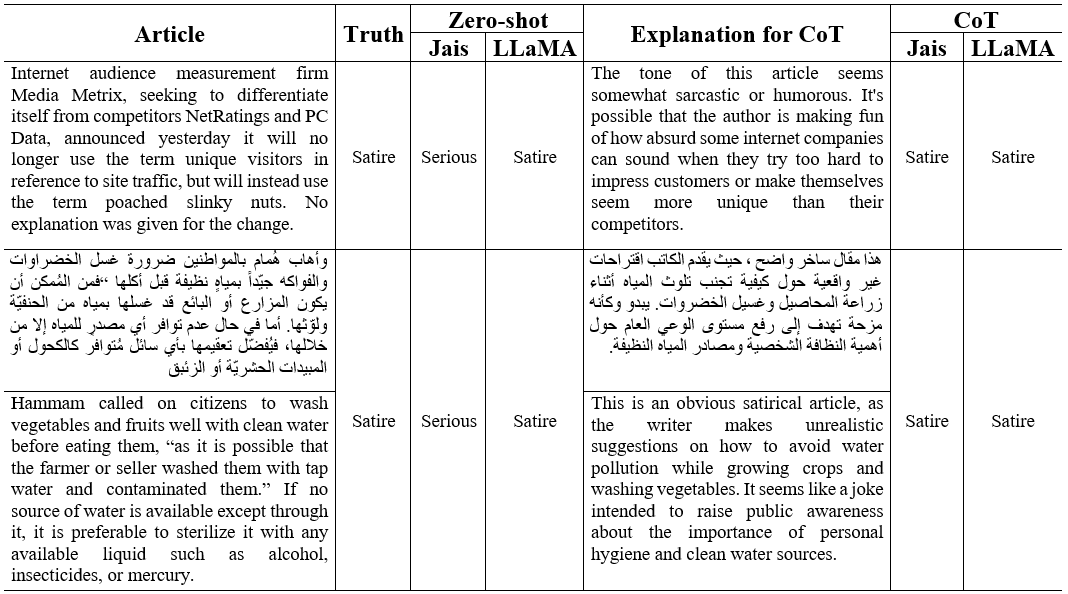}
  \caption{Examples of CoT \& Zero-Shot}
  \label{fig:sample}
\end{figure*}

% We collected the datasets used in our study from the respective papers published in ACL Anthology\footnote{\href{https://aclanthology.org/}{https://aclanthology.org/}}. 
The summary of the datasets is shown in Table~\ref{tab:dataset_stats}. The first dataset is ``Assiri''~\cite{app131910616}, an Arabic dataset that encompasses 760 satirical articles and 765 non-satirical articles. The ``Saadany''~\cite{saadany2020fake} is an Arabic dataset that, originally, comprises 3185 satirical articles. To balance the dataset, we merged it with the ``bbc-arabic-utf8'' dataset from ``SourceForge''\footnote{\href{https://sourceforge.net/}{https://sourceforge.net/}} website, comprising of 4763 non-satirical articles. The ``Phosseini'' dataset~\cite{li-etal-2020-multi-modal} is an English dataset comprising of 3956 satirical articles and 2987 non-satirical articles. The ``SatiricLR'' dataset~\cite{frain-wubben-2016-satiriclr} is an English dataset that encompasses 1706 satirical articles and 1705 non-satirical articles.
% \newpage
% Dataset description: Cite all datasets. Give their statistics as a table.

\section{Results}\label{sec:results}
% \textcolor{blue}{We conducted experiments using two models, Jais-chat and LLaMA-2-chat, on four different datasets in English and Arabic. The performance of the models was evaluated using accuracy, precision, recall, and F1-score. The experimental results are summarized in Table~\ref{tab:results}.}

As observed in Table~\ref{tab:results}, the Jais-chat model exhibits superior performance when utilizing the CoT prompting approach compared to zero-shot prompting across all scenarios. Jais-chat achieves its highest F1-score of 80\% with English prompts using CoT prompting, outperforming its performance with the Arabic prompts, where the highest F1-score is 70\%, respectively. In contrast, the LLaMA-2-chat model shows minimal improvements with the CoT approach compared to the zero-shot approach, with F1-scores reaching 72.5\% for English prompts and 73\% for Arabic prompts, respectively. This indicates that while CoT prompting significantly benefits the Jais-chat model, the LLaMA-2-chat model performance remains relatively consistent, when prompted with zero-shot and CoT. This observation indicates that LLaMA-2-chat is not tuned specifically for CoT prompting and hence showed same performance regardless of the prompting strategy. A sample article is provided in Figure~\ref{fig:sample} along with the ground truth and predictions for both models, Jais-chat and LLaMA-2-chat, when prompted with zero-shot and CoT. (For convenience, the Arabic text has been translated.)

It is worth noting that the LLaMA-2-chat model achieved exceptional recall scores across all datasets, exceeding 97\%. This suggests that while the model may struggle with precision, it is highly effective at identifying relevant instances, potentially indicating a tendency to classify more instances as positive. Over-classifying instances as satirical risks dismissing legitimate information, while over-classifying instances as non-satirical could lead to the spread of false information as credible. Both scenarios contribute to the spread of misinformation. Therefore, the trade-off between recall and precision should be carefully considered in the context of satire detection.

% Must contain one main table, and one additional table (please come with the idea on your own).
\section{Conclusion}\label{sec:conclusion}
This study explores the efficacy of satire detection using multilingual models utilizing different prompting techniques, comparing the bilingual Jais-chat model with the multilingual LLaMA-2-chat model. Referring to the research questions, we observe that the multilingual LLaMA-2-chat model produces consistently stable outcomes regardless of the prompting technique. In contrast, the bilingual Jais-chat model demonstrates more variable results, showing significantly improved performance with CoT prompting compared to zero-shot prompting. %We focused on two primary approaches: zero-shot prompting and chain-of-thought (CoT) prompting, assessing their performance on satire detection in both English and Arabic datasets. 
The results indicate that CoT prompting improves or maintains performance depending on the model.

Future work should aim to refine these models, expand datasets, and include more languages to better address the complexities of satire in diverse cultural contexts. Improving satire detection methodologies can enhance public understanding of media content and reduce the spread of misinformation in an increasingly complex information landscape.

% The study contributes to the literature on satire detection by demonstrating the advantages of CoT prompting and highlighting the need for further exploration of multilingual capabilities in current detection methods. 
% Summary of the paper.

% If you had one more month, what more could you do?

% If you had six more months, what more could you do?

\section*{Ethical Considerations}
Satire detection in multilingual contexts presents important ethical challenges. One key concern is misclassifying satire as misinformation or the reverse, especially when cultural nuances are overlooked. This can unintentionally spread misinformation or diminish legitimate satire. Bias in large models like Jais-chat and LLaMA-2-chat is another issue. Since humor varies greatly across cultures, these models may reinforce harmful stereotypes or misinterpret satire, particularly if the training data lacks diversity. Ultimately, it is crucial to deploy satire detection models carefully, ensuring transparency and minimizing potential negative impacts on public discourse.

\section*{Limitations}
% This research has a few limitations. First, the models' effectiveness relies heavily on the quality of the prompts. While Chain-of-Thought (CoT) prompting can improve results, poorly designed prompts may lead to unreliable outcomes. Second, the research study should be extended to include languages beyond just English and Arabic to ensure broader applicability and validate its findings across diverse linguistic contexts. Third, the datasets mainly consist of written satire, which may hinder the models' ability to detect satire in other formats like images or videos. Finally, the study emphasizes full news articles, neglecting shorter formats such as headlines or social media posts. Future work should investigate model performance across different text lengths and types of satire.
This research has several limitations. First, the effectiveness of both Jais-13b-chat and LLaMA-2-chat models relies heavily on the quality of prompts, and while Chain-of-Thought (CoT) prompting can enhance results, poorly designed prompts may yield unreliable outcomes. Additionally, our study focuses solely on English and Arabic, limiting the generalizability of our findings to other linguistic contexts; future research could address this by incorporating additional languages to validate applicability across a broader spectrum. Another limitation is that our datasets predominantly contain written satire, potentially reducing the models' ability to detect satire in multimedia formats such as images or videos. Furthermore, our analysis centers on full news articles, omitting shorter forms of satire, such as headlines and social media posts. Lastly, the differences between Jais-13b-chat and LLaMA-2-chat extend beyond the bilingual versus multilingual training scope, including variations in model architecture and fine-tuning strategies, which prevent a pure comparison based on language coverage alone. Future work should explore model performance across diverse text formats, lengths, and controlled conditions isolating language-focused training differences.

% List the limitations of your work here. Is it dependent on the prompt? Do you consider full news article? Etc.

\bibliography{anthology}
\bibliographystyle{acl_natbib}
% \section*{Appendix}
% \appendix{appendix}    
\end{document}